%% file: main.tex
\documentclass[10pt,twocolumn,letterpaper]{article}
\usepackage{multirow}

\usepackage[pagenumbers]{cvpr} 
\input{preamble}

\definecolor{cvprblue}{rgb}{0.21,0.49,0.74}
\usepackage[pagebackref,breaklinks,colorlinks,citecolor=cvprblue]{hyperref}

\usepackage{color, colortbl}
\definecolor{lgray}{rgb}{0.88,0.88,0.88}

\def\paperID{?} 
\def\confName{CVPR}
\def\confYear{2024}

\title{oTTC: Object Time-to-Contact for Motion Estimation in Autonomous Driving}

\author{Abdul Hannan Khan$^{1,2}$, Syed Tahseen Raza Rizvi$^{2}$, Dheeraj Varma Chittari Macharavtu$^{1}$, Andreas Dengel$^{1,2}$\\
Department of Computer Science, RPTU Kaiserslautern-Landau$^{1}$,\\
German Research Center for Artificial Intelligence (DFKI GmbH)$^{2}$,\\
67663 Kaiserslautern, Germany\\
{\tt\small Corresponding Author: hannan.khan@dfki.de}
}

\begin{document}
\maketitle
\input{sec/0_abstract}    
\input{sec/1_intro}
\input{sec/2_related}
\input{sec/3_gt_gen}
\input{sec/5_ottc}
\input{sec/4_eval_set}
\input{sec/6_results}
\input{sec/7_conclude}
{
    \small
    \bibliographystyle{ieeenat_fullname}
    \bibliography{main}
}


\end{document}

%% file: preamble.tex
\usepackage[dvipsnames]{xcolor}


%% file: sec/0_abstract.tex
\begin{abstract}
Autonomous driving systems require a quick and robust perception of the nearby environment to carry out their routines effectively. With the aim to avoid collisions and drive safely, autonomous driving systems rely heavily on object detection. However, 2D object detections alone are insufficient; more information, such as relative velocity and distance, is required for safer planning. Monocular 3D object detectors try to solve this problem by directly predicting 3D bounding boxes and object velocities given a camera image. Recent research estimates time-to-contact in a per-pixel manner and suggests that it is a more effective measure, than velocity and depth combined. However, per-pixel time-to-contact requires object detection to serve its purpose effectively and hence increases overall computational requirements as two different models need to run. To address this issue, we propose per-object time-to-contact estimation by extending object detection models to additionally predict the time-to-contact attribute for each object. We compare our proposed approach with existing time-to-contact methods and provide benchmarking results on well-known datasets. Our proposed approach achieves higher precision compared to prior art while using a single image.
\end{abstract}

%% file: sec/1_intro.tex
\section{Introduction}
\label{sec:intro}

\begin{figure}[!t]
    \centering
    \includegraphics[width=0.46\textwidth]{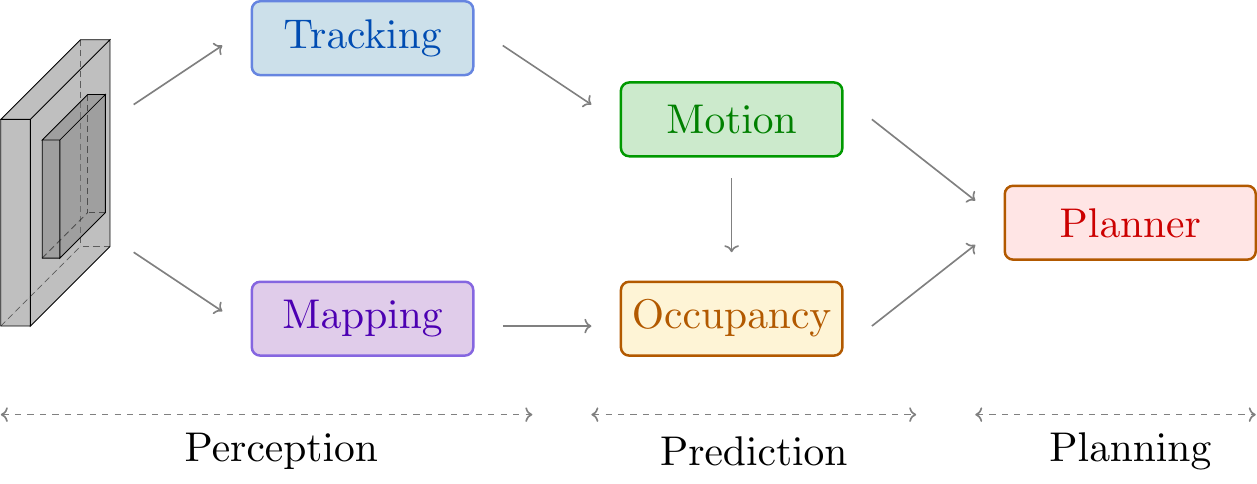}
    \caption{An autonomous driving pipeline including perception, prediction and planning submodules; inspired by planning oriented autonomous driving \cite{hu2023planning}.}
    \label{fig:ad_pipeline}
\end{figure}

Autonomous driving aims to improve urban traffic and ensure passenger safety. Avoiding collisions and planning safe maneuvers are core tasks of autonomous driving systems. The existing pipelines for such systems contain redundancy, making them slow and computationally expensive. These issues negatively affect the performance of autonomous driving systems. Fig. \ref{fig:ad_pipeline} shows an autonomous driving pipeline, divided into multiple modules, i.e., perception, prediction and planning. Motion and occupancy form the prediction module, where occupancy prediction focuses on the static part of the scene, such as the drivable area and lanes, while motion prediction handles the moving objects or the dynamic part of the scene.

Further, existing motion prediction approaches model motion as position and velocity \cite{hu2023planning}; however, modeling the motion with these absolute values is unnecessary. Since occupancy prediction already handles the drivable area and obstacles, motion prediction should only provide a risk metric for each object, which helps the planning module control the vehicle's speed and braking. Even when a risk metric is computed using position and velocity, its precision is lower than the precision of either of its estimated components. Conclusively, in autonomous driving, modeling motion using velocity and position is suboptimal.

Badki et al., \cite{badki2021binary} emphasize that in driving scenarios, time-to-contact (TTC) is more effective than velocity and depth of an object to assess its risk. Also, TTC can be computed using the ratio of depths of an object in two consecutive frames, or depth and velocity. The resultant TTC value is a ratio, while the velocity and depth are metric values; hence, it is easier to predict TTC even when predicting depth or velocity is ill-posed \cite{badki2021binary}. However, little research exists on TTC prediction \cite{badki2021binary, hur2020self, yang2020upgrading} compared to velocity \cite{tian2019fcos, wang2021fcos3d} and depth estimation \cite{patil2022p3depth, lee2019big, chang2019deep, tankovich2021hitnet}.

\begin{figure}[!t]
\centering
\subfloat[][]{\includegraphics[width=0.115\textwidth]{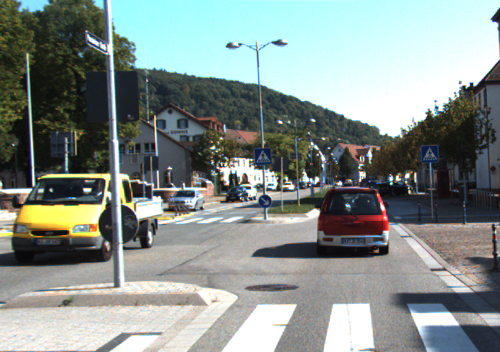} \label{fig:comp_o}}
\subfloat[][]{\includegraphics[width=0.115\textwidth]{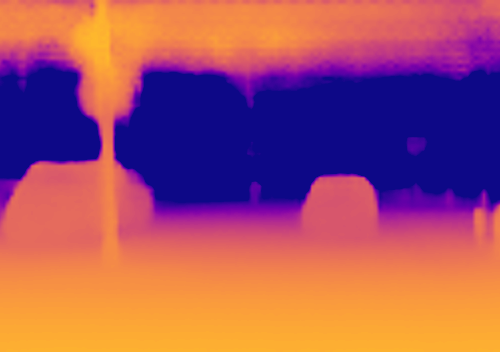} \label{fig:comp_d}}
\subfloat[][]{\includegraphics[width=0.115\textwidth]{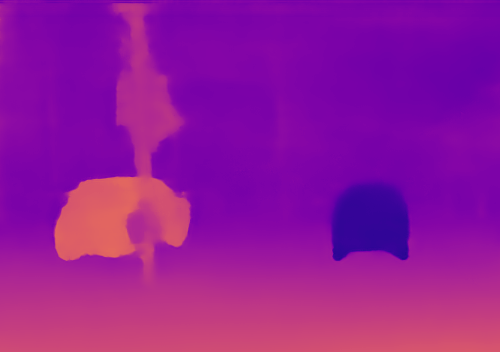} \label{fig:comp_c}}
\subfloat[][]{\includegraphics[width=0.115\textwidth]{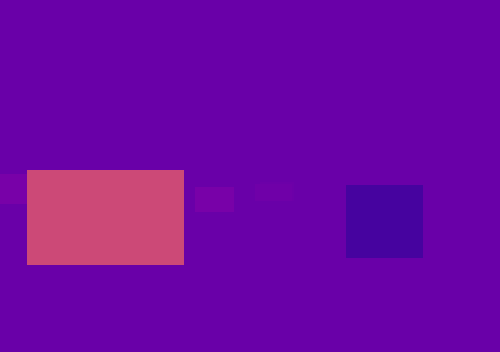} \label{fig:comp_t}}

\caption{Shows monocular depth predictions (b), per-pixel time-to-contact predictions (c) and oTTC predictions (d). (b) The depth prediction shows how far the objects are from the camera, with hotter meaning closer to the camera. (c) The per-pixel TTC predictions show the relative motion of pixels from ego perspective, where temperature indicates how fast objects are moving toward the ego vehicle. The depth prediction (b) is based on PixelFormer \cite{agarwal2023attention} and per-pixel time-to-contact prediction (c) is based on Binary TTC \cite{badki2021binary}. (d) is the output of our proposed oTTC model, where normal temperature shows background while hot and cold show objects moving towards and away, respectively, from ego vehicle.
}
\label{fig:comp}
\end{figure}

Binary TTC \cite{badki2021binary} estimates TTC in a per-pixel fashion. While per-pixel TTC heatmaps provide precise class-agnostic information about obstacles, without object detections, they can create false alarms, i.e., the TTC heatmaps show low TTC values for pixels that belong to the road but are next to the ego vehicle, as shown in Fig. \ref{fig:comp} (c). This makes TTC estimation dependent on object detection to estimate risk, despite both being two separate tasks with individual computational costs. Furthermore, TTC heatmaps include TTC values for pixels belonging to background classes, which are not entirely relevant, i.e., TTC values for the sky, treetops, roads, skyscrapers, etc.

To avoid the additional computational costs and estimate precise risk per object, we propose to predict TTC as an object attribute using a center detection and attribute prediction strategy \cite{liu2019high}. This is done by adding an extra branch to predict TTC along with other attribute prediction branches, i.e., scale and offset. Following the approach in \cite{badki2021binary}, we predict the ratio of depths instead of TTC, as it is linear from the camera perspective. The contributions of this paper are multifold:

\begin{itemize}
    \item We explore the methods to generate TTC ground truth using existing object detection and tracking datasets.
    \item We extend 2D object detection network to predict the binary and continuous time-to-contact per object.
    \item We evaluate our model on a TTC estimation benchmark, where it beats the SOTA with a significant margin.
    \item We benchmark proposed approaches on multiple autonomous driving datasets, both real and synthetic.
    \item We present explainability analysis on proposed approaches to realize important features for TTC prediction.
\end{itemize}

%% file: sec/2_related.tex
\section{Related Work}
\label{sec:related_work}

Autonomous vehicles require a rich 3D environment perception to avoid collision with obstacles and safely manoeuvre in drivable space. However, expensive and complex additional sensors are required to acquire 3D data from the environment, which further needs to go through computationally expensive processing to transform it into a useful form. This setup renders the approach impractical for real-time and cost-effective scenarios.

\subsection{Time-to-Contact Estimation}

Time-to-Contact is well studied in psychophysics and proven to be more effective indicator for a driver compared to depth and velocity \citep{lee1976theory}. Time-to-Contact can be estimated directly from images while estimating depth and velocity is an ill-posed problem, as TTC depends on their ratios \citep{horn2009hierarchical}. \citep{meyer1992estimation, meyer1994time, camus1995calculating} estimate TTC from optical flow. \citep{horn2007time} use constant brightness assumption to estimate TTC directly, without the requirement of estimating optical flow first, however, these method requires masks of the object of interest to be effective.

Recent work by Badki et al. \citep{badki2021binary}, estimates TTC in a per pixel fashion, however, this method is not only dependent on further information like object detections to be fully applicable, but also estimates TTC for background pixel, which is unnecessary. We take a rather objective approach by modelling TTC as an object attribute, which can be easily estimated along with corresponding object detections with negligible computational overhead.

\subsection{Object Detection and Attribute Prediction}
Ross Girshick et al. \citep{girshick2014rich} proposed R-CNNs as an initial bridge between object classification and object detection. Recent R-CNN based techniques \citep{cai2018cascade, he2017mask, liu2021swin, girshick2015fast, ren2015faster} have improved a lot on the basic idea of R-CNNs by contributing towards both efficiency and performance. To eliminate region proposal network from R-CNNs and hence achieve higher efficiency YOLO, SSD and RetinaNet were proposed \citep{redmon2016you, liu2016ssd, lin2017focal} resulting in single stage architectures. However, these architectures focus on efficiency and compromise on performance to achieve it.

Recently, anchor-free object detection architectures \citep{duan2019centernet, law2018cornernet, tian2019fcos, khan2022f2dnet, khan2023localized} were proposed to bridge the gap between performance and efficiency and perform object detection in end to end fashion exploiting the power of deep convolutional networks. CornerNet \citep{law2018cornernet} proposes to detect an object as a pair of key-points, bounding box corners in the case of 2D detection, however the approach can be easily extended to pose estimation. CenterNet \citep{duan2019centernet} models objects as triplet instead of paired key-points where corner points are used for bounding box proposal generation and center point is used for verification. FCOS \citep{tian2019fcos} predicts objects in per-pixel fashion. FCOS3D \citep{wang2021fcos3d} extends 2D object detection to mono-3D object detection and predicts additional object attributes like velocity and orientation angle per object.

We follow the approach of attribute prediction along with 2D object detection to predict TTC per object \cite{tian2019fcos, wang2021fcos3d}. The approach uses a single camera image as an input, and yet outperforms approaches with stereo or sequence input.

%% file: sec/3_gt_gen.tex
\section{Ground Truth Data Generation}
Deep Learning models heavily rely on a substantial amount of data; however, there is currently no  well-established dataset that contains TTC ground truths. However, it is possible to calculate TTC ground truth values using different object and scene properties. In this section, we discuss various methods to generate TTC ground truth values from the depth, velocity, and tracks of the objects.

\subsection{Depth and Velocity to TTC}
A simpler way to calculate TTC per object is to use its depth and velocity. Eq. \ref{eq:depth_vel} shows the relation between TTC $\tau$, depth and the velocity.

\begin{equation}
    \tau = \frac{\lambda}{\lambda^\prime} \:,
    \label{eq:depth_vel}
\end{equation}

where $\lambda$ indicates the depth of the object and $\lambda^\prime$ indicates the velocity of the object.

It is important to note that the depth here is the object distance from the camera center, rather than $Z$ of the 3D object position $(X, Y, Z)$. Using $Z$ and $Z^\prime$ generates imprecise TTC values as lateral movement is ignored, which can be important in the case of objects that move laterally, like pedestrians at cross-walks and vehicles at junctions

Further, the depth and velocity annotations require information from additional sensors like LIDAR and RADAR. This makes, TTC from depth and velocity expensive and prone to precision errors.

\subsection{3D Object Tracks to TTC}
\label{sec:gt_3dot}

3D object tracking datasets contain object annotations with 3D locations, as well as their correspondences between consecutive frames. The depth values of the object in consecutive frames can be used to calculate TTC using the following relations;

\begin{equation}
    \eta = \frac{\lambda(t_{1})}{\lambda(t_{0})},
    \label{eq:rod}
\end{equation}

where $\lambda$ represents the distance of the object from the camera center \cite{yang2020upgrading}. The motion-in-depth $\eta$ is then converted into TTC using:

\begin{equation}
    \tau = \frac{T}{1 - \eta},
    \label{eq:ttc}
\end{equation}

where $T$ is the time delta between two consecutive key frames \cite{yang2020upgrading}, which is also the multiplicative inverse of key frames per second of the video sequence ($kfps$), here key frame indicates the frame with given ground truth.

However, the problem of additional sensor requirements remains, as 3D object tracking annotations require depth information of the scene.

\subsection{2D Object Tracks to TTC}

Assuming rigid, planar and fronto-parallel objects, the perceived height of the object is given by;

\begin{equation}
    h = f\frac{H}{\lambda},
    \label{eq:perceived_height}
\end{equation}

where $f$ is the focal length of the camera, $H$ is the metric height of the object, and $\lambda$ is the distance of the object from the camera.

Using Eq. \ref{eq:perceived_height} in Eq. \ref{eq:rod} gives;

\begin{equation}
    \eta = \frac{\lambda(t_{1})}{\lambda(t_{0})} = \frac{h(t_{0})}{h(t_{1})}.
    \label{eq:roh}
\end{equation}

However, the planer and fronto-parallel assumption is an oversimplification of the problem. Due to multiple road lanes and intersections, a higher percentage of traffic objects are cars, which are neither planar nor always fronto-parallel. Moreover, the Eq. \ref{eq:perceived_height} and \ref{eq:roh} holds if the orientation of the vehicle is either parallel or perpendicular to the camera axis, as at these angles the bounding box height is equal to the height projected on the camera plane.

Furthermore, the projected height of the object ($h_p$) can be calculated from the bounding box height given the projected orientation angle using the following relation;

\begin{equation}
    h_p = \frac{h_{bbox} \cos(\theta) - w_{bbox} \sin(\theta)}{\cos^{2}(\theta) - \sin^{2}(\theta)},
    \label{eq:ph2h}
\end{equation}

where $h_{bbox}$ and $w_{bbox}$ are the height and width of the bounding box, and $\theta$ is the projected orientation angle. The above relation, when used with Eq. \ref{eq:roh} can generate accurate $\eta$ values as long as the projected orientation angle remains constant between two consecutive frames. While this condition might not always apply, the error due to change in projected orientation angle can be minimized by using higher $fps$, which results in smaller change in projected orientation angle.

%% file: sec/5_ottc.tex
\begin{figure*}[!t]
    \centering
    \includegraphics[width=0.98\textwidth]{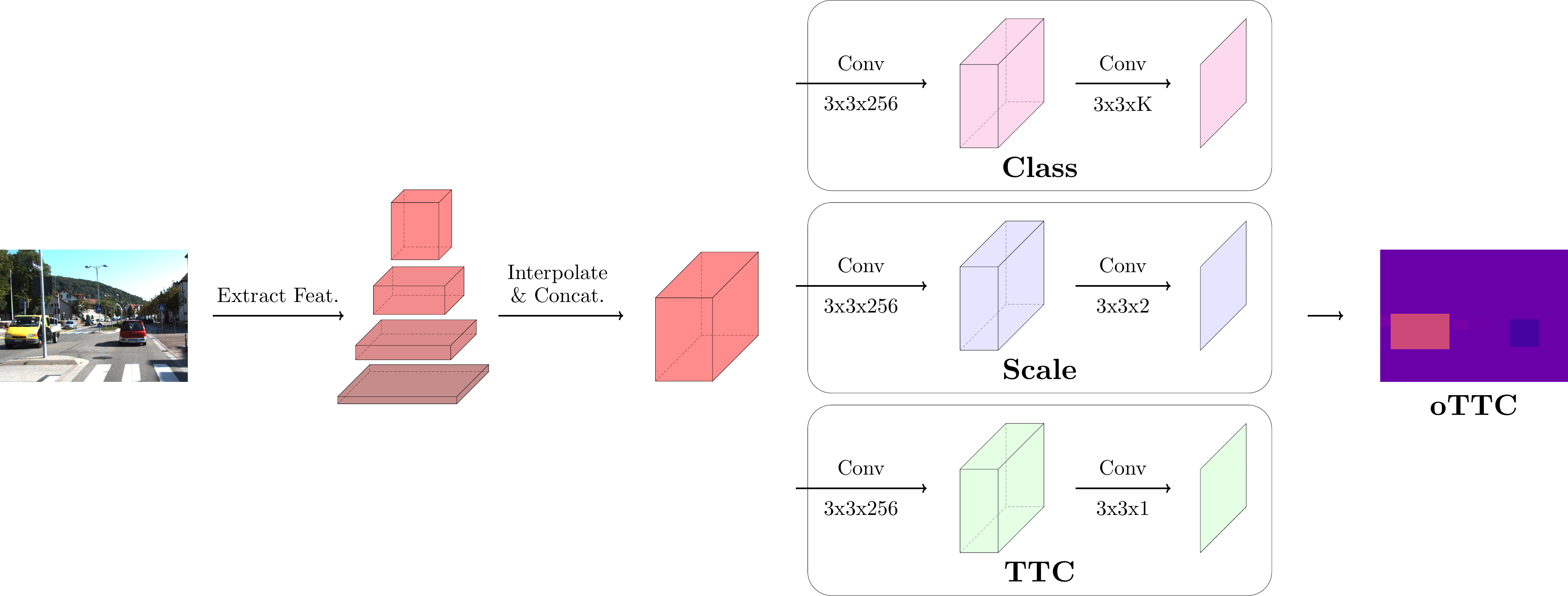}
    \caption{Simplified architecture of our oTTC model. It uses HRNetW32 \cite{wang2020deep} backbone to extract feature. Some details have been omitted for better visibility and to convey general idea of the architecture.}
    \label{fig:arc}
\end{figure*}

\section{Object Time-to-Contact}
Due to the camera motion and the dynamic nature of the traffic scene, the captured images are subject to motion blur. The amount of motion blur is directly proportional to the relative motion of the objects and the camera's exposure time. With a constant or known exposure time, the motion-in-depth (MiD) depends solely on motion blur. Modern deep-learning techniques can estimate motion blur as well as exposure time, given a large amount of data, and exploit this information to estimate MiD. The MiD values can be converted to time-to-contact (TTC) using Eq. \ref{eq:ttc}, given the capture frequency.

The 2D object detectors predict location information for each of the objects in the scene. The location information itself doesn't convey any reliable clues regarding the TTC of the object due to perspective distortion and motion ambiguity. With the aim of exploiting the hidden motion information inside motion blur, we try to directly estimate MiD per detected object. For this purpose, we use CSP \cite{liu2019high} as the base and add an extra branch in the detection head to predict TTC. 

The input image is passed through HRNetW32 \cite{wang2020deep} to extract features. These features of different scales are then interpolated to obtain uniform-sized feature maps. These feature maps are then concatenated and passed through convolution blocks. The final enriched feature volume goes in parallel through class, scale, offset and TTC branches. The class, scale and offset are then combined to create bounding box predictions along with the object classes. Finally, the object detections and TTC predictions are combined to generate an oTTC map. Fig. \ref{fig:arc} shows our proposed architecture; the convolution block after interpolation and concatenation is skipped in the diagram along with the offset branch for simplicity.

Time-to-contact values vary from $-\infty$ to $+\infty$. Directly predicting such large values can result in underfitting given a relatively small amount of data. Therefore, to avoid dealing with a large range of values, we follow \cite{badki2021binary} and predict MiD $\eta$ instead and calculate TTC afterward using Eq. \ref{eq:ttc}. Moreover, we use L1Loss to optimize the TTC branch with $1\times10^{-2}$, $5\times10^{-2}$, $1\times10^{-1}$ and $2\times10^{-1}$ as loss weights for class, scale, offset and TTC branch, respectively.



%% file: sec/4_eval_set.tex
\section{Data Preparation and Evaluation Settings}
This section includes the details of datasets, data preparation, evaluation criteria, and, at the end, the training and testing environment used for experiments.

\subsection{Datasets}
To evaluate the TTC prediction approaches, we use well-known object tracking datasets, including real and synthetic datasets. The tracking datasets include object tracks, which are necessary to obtain time-to-contact ground truths. NuScenes is a 3D dataset which contains $1000$ scenes, recorded in Boston and Singapore, comprising $1.4$ million object annotations \cite{caesar2020nuscenes}. Shift dataset \cite{sun2022shift} is a synthetic dataset containing $2.5$ million images, including both day and night scenes. The KITTI object tracking dataset \cite{geiger2013vision} contains both 2D and 3D annotations comprising $50$ video sequences with objects belonging to $8$ classes. However, we only benchmark \textit{Car} and \textit{Pedestrian} classes.

\subsection{Data Preparation}
Both the KITTI tracking \cite{geiger2013vision} and NuScenes \cite{caesar2020nuscenes} datasets do not contain TTC information in ground truth annotations. However, this information can be easily extracted using object tracks and location information. We extract motion-in-depth (MiD) \cite{yang2020upgrading} of the object from its depth values in consecutive key frames. In the case of the KITTI tracking  \cite{geiger2013vision} key-frames per second ($kfps$) is $10$ and in the case of the NuScenes dataset, $kfps$ is $2$ \cite{caesar2020nuscenes}.

The NuScenes dataset \cite{caesar2020nuscenes} does not contain 2D detection annotations; we generate 2D bounding boxes by projecting 3D bounding box points into the camera image. The resultant bounding boxes are not tight; however, they serve the purpose, as we do not aim to evaluate object detections.

\subsection{Evaluation Metric}
A evaluation metric provides a quantitative measure to compare the performance of different solution to a single problem. We use two different metrics to evaluate our approaches. We use MiD in all of our experiments except for binary risk prediction where we use risk accuracy. 

\subsubsection{Motion-in-Depth}

The MiD \cite{yang2020upgrading} $\eta $ represents the proportional change in depth of the object, with values mostly lying between $(0.5,1.3)$, where $\eta < 1$ represents the object moving toward the ego vehicle and $\eta > 1$ represents the object moving away from the ego vehicle. MiD-Loss is commonly used to evaluate continuous time-to-contact \cite{badki2021binary} \cite{yang2020upgrading}; it is given by:

\begin{equation}
    MiD \, Loss = \lVert \log(\eta) - \log(\eta_{GT}) \rVert_1 \times 10^4,
    \label{eq:mid}
\end{equation}

where $\eta_{GT}$ and $\eta$ represent ground truth and predicted MiD-values. We calculate $oMiD$, which considers $MiD$ per object instead of per pixel. We also calculate $oMiD^+$ which only considers values $\eta < 1$. $oMiD^+$ focuses on objects moving toward the ego vehicle, which are more risky.

\subsubsection{Binary Risk Accuracy}
Besides continuous TTC, we also experiment with binary risk prediction, where the model predicts whether an object is risky or not, i.e. $\eta < 1$. To evaluate binary risk prediction, we use accuracy as the metric, defined as;

\begin{equation}
    Accuracy = \frac{TP}{FP + TP + FN},
\end{equation}

where $TP$ is true positive, $FP$ is false positive and $FN$ is false negative binary risk prediction.

\subsection{Training and Evaluation Settings}
\label{subsec:eval_set}
For TTC, we perform evaluation in a detection agnostic way so that it is not affected by the detection performance of the network. To do so, we take object centers from object detection ground-truth, instead of predicted object centers, and use them to extract predicted TTC values from the predicted TTC maps.


The KITTI tracking dataset \cite{geiger2013vision} includes $50$ videos in total, with $21$ for training and $29$ for testing purposes. As testing sequences are withheld, i.e., the annotations for these sequences are not provided and testing servers do not include time-to-contact benchmarks, we randomly take $4^{th}$ and $11^{th}$ video sequence from the train set for evaluation.

For continuous TTC evaluation, existing methods use KITTI scene flow validation set, containing $40$ samples. As the proposed method for continuous TTC predicts TTC per object while the existing research predicts it in a per pixel manner, it is not possible to directly compare them. However, there is an overlap between the KITTI tracking dataset and scene flow validation set, containing $19$ out of $40$ samples. We evaluate our continuous TTC method on these $19$ samples, just to establish a ground to evaluate the performance of the proposed method. For binary risk estimation, we use the validation set of KITTI tracking dataset, ignore annotations with $0.998 \leq \eta < 1.002$, and consider $\eta < 0.998$ as risky.

Training is performed on $4$ A100 GPUs, with $8$ images per GPU. And the evaluation is performed in one image per batch manner on a single V100 GPU.

%% file: sec/6_results.tex
   \begin{table}[!t]
        \renewcommand{\arraystretch}{1.3}
        \caption{Time-to-contact benchmark on KITTI dataset \cite{geiger2013vision}. (SF) indicates that the evaluation based on scene flow validation set while (T) indicates the evaluations based on video sequence $4$ and $11$ of KITTI tracking dataset \cite{geiger2013vision}. * indicates that the evaluation is based on 19 samples out of 40 samples of scene flow validation set, as described in Sec. \ref{subsec:eval_set}.}
        \label{tab:benchmark_kitti}
        \centering
        \resizebox{\linewidth}{!}{
        \begin{tabular}{llcc}
            \hline
             Method & Setup & $oMiD$ (SF) & $oMiD$ $|$ $oMiD^{+}$ (T) 
             \\
             \hline
             PRSM \cite{vogel20153d} & Stereo & 124 &  -  \\
             OSF \cite{menze2015object} & Stereo & 115 & -    \\
             \hline
             Hur \& Roth \cite{hur2020self} & Sequence & 115 & -   \\
             Yang \& Ramanan \cite{yang2020upgrading} & Sequence & 75 & 118 $|$ 127 
             \\
            Binary TTC \cite{badki2021binary} & Sequence &  74 & 120 $|$ 124 
            \\
             \hline
             oTTC (ours) & Monocular & \textbf{*44} & \textbf{107} $|$ \textbf{102} 
             \\
             \hline
        \end{tabular}
        }
    \end{table}


\section{Experiments \& Results}

\begin{table}[!t]
        \renewcommand{\arraystretch}{1.3}
        \caption{Binary risk prediction results on autonomous driving datasets. Risk accuracy is higher on real datasets, i.e., NuScenes \cite{caesar2020nuscenes} and KITTI \cite{geiger2013vision}, compared to Shift dataset which is synthetic.
        }
        \label{tab:brp}
        \centering
        \resizebox{\linewidth}{!}{
        \begin{tabular}{lccc c ccc}
            \hline
              & \multicolumn{3}{c}{Risk Accuracy} && \multicolumn{3}{c}{mAP} 
             \\
             \cline{2-4}
             \cline{6-8}
             Dataset & All & Pedestrian & Car && All & Pedestrian & Car \\
             \hline
             NuScenes & 89.9 & 93.5 & 88.5 && 22.1 & 16.9 & 27.3
             \\
             KITTI & 91.2 & 99.6 & 90.5 && 40.8 & 22.8 & 65.7
             \\
             Shift & 83.3 & 90.6 & 79.1 && 39.6 & 34.3 & 39.5
             \\
             \hline
        \end{tabular}
    }
    \end{table}

Although, predicting TTC compared to depth and velocity is easier, predicting TTC per object using a single image is still an ill-posed problem. The understanding of traffic scene can help to determine the direction of travel, but not the absolute or relative speed. However, captured camera motion artifacts, like motion blur, contain potential clues that can determine if an object is moving towards or away from the camera along with the rate of this change upto a certain precision. To verify this and test how much information can be extracted from these motion artifacts, we conduct two experiments. First, we perform continuous TTC prediction per object and compare our results with existing state-of-the-art. 
Second, we try binary risk prediction where we classify objects as risky and non-risky based on if they are moving towards the camera, i.e., $TTC > 0$. We also compare per object TTC prediction with TTC calculated from predicted velocity and depth to establish which has better precision. Finally, we compare per object TTC prediction with TTC calculated from tracked depth values using mono3D object tracking \cite{wang2021fcos3d}.

\begin{table}[!t]
        \renewcommand{\arraystretch}{1.3}
        \caption{Comparison of oTTC with FCOS3D \cite{wang2021fcos3d} on NuScenes dataset \cite{caesar2020nuscenes}. The $oMiD$ values for FCOS3D are calculated using velocity and depth predictions. oTTC beats FCOS3D in TTC prediction, which shows that the direct TTC prediction is better than TTC calculated from depth and velocity predictions.}
        \label{tab:mono3d_nu}
        \centering
        \resizebox{0.95\linewidth}{!}{
        \begin{tabular}{lcccccc}
            \hline
             Method / $oMiD$ & Car & Truck & Bus & Ped. & Mot. & All \\
             \hline
             FCOS3D & 415 & 439 & \textbf{434} & \textbf{184} & 482 & 391 
             \\ 
             oTTC (ours) & \textbf{351} & \textbf{394} & 468 & 291 & \textbf{376} & \textbf{376} 
             \\
             \hline
        \end{tabular}
        }
    \end{table}

\subsection{Object Time-to-Contact (oTTC)}

To benchmark our oTTC approach, we compare it against existing approaches on Kitti dataset \cite{geiger2013vision} as it is the only dataset with TTC benchmarks available. Table \ref{tab:benchmark_kitti} shows the benchmarks on Kitti dataset. $MiD$ represents per pixel motion-in-depth, while $oMiD$ and $oMiD^+$ represent motion-in-depth per object. Binary TTC \cite{badki2021binary} predicts TTC in a per pixel fashion which is not directly comparable to our approach, to make it comparable, we take Binary TTC \cite{badki2021binary} outputs on Kitti dataset and extract values of $MiD$ at centre pixels of the objects. It is evident that, our oTTC, while being efficient, works better compared to per pixel TTC prediction. $oMiD^+$ shows $oMiD$ values of the objects moving toward the ego vehicle, and it is evident that, proposed approach is even better when considering $oMiD^+$. Moreover, our approach uses a single image only, while the rest of the approaches use multiple images.

\begin{table}[!t]
        \renewcommand{\arraystretch}{1.3}
        \caption{Comparison of oTTC with Center Track \cite{zhou2020tracking} on KITTI dataset \cite{geiger2013vision}. oTTC has better $oMiD$ and $oMiD^+$, which indicates that direct TTC prediction yields precise values compared to TTC calculated from predicted depth values.}
        \label{tab:track_kitti}
        \centering
        \resizebox{0.8\linewidth}{!}{
        \begin{tabular}{llcc}
            \hline
             Method & Input & $oMiD$ & $oMiD^{+}$ 
             \\
             \hline
             Center Track & Two Images & 116 & 108 
             \\
             oTTC (ours) & Single Image & 107 & 102 
             \\
             \hline
        \end{tabular}
        }
    \end{table}

\subsection{Binary Risk Prediction}
In binary risk prediction, we try to predict if the traffic object is moving towards or away from the ego vehicle, based on a single image. Although, the vehicles on the apposite sides of a two-way road and relatively static objects like pedestrians move mostly towards the ego vehicle, the objects in the same lane or the lanes next are more important. These objects can highly impact the decisions as they are closer to the ego vehicle and can move both towards and away from the ego vehicle, hence posing greater risk. To this extent, we conduct binary risk prediction experiment by predicting if $TTC > 0$ for each object. To establish generalization, we use multiple, real and synthetic datasets. Also, similar to continuous TTC prediction, we use CSP \cite{liu2019high} as detection architecture and predict binary risk estimates instead of continuous TTC per object.

Tab. \ref{tab:brp} shows detailed results of binary risk prediction. It is evident that, binary risk prediction works better for real datasets, i.e., NuScenes \cite{caesar2020nuscenes} and KITTI \cite{geiger2013vision} compared to Shift \cite{sun2022shift} dataset which is synthetic. This shows that, motion artifacts, which are present in real datasets but not synthetic datasets, can aid in TTC based risk estimation.

\begin{figure*}[!t]
\centering
\addtolength{\tabcolsep}{-5pt}
\renewcommand{\arraystretch}{0.4}
\begin{tabular}{ccccccc}
    & {\small \textit{Image}} & {\small \textit{GT}} & {\small \textit{oTTC}} & {\small \textit{Intp. bTTC}} & {\small \textit{Intp. oTTC}}
    \vspace{0.3em}\\
     \raisebox{2.6em}{\rotatebox[origin=t]{90}{{\small \textit{Moving}}}}& \includegraphics[width=0.23\textwidth]{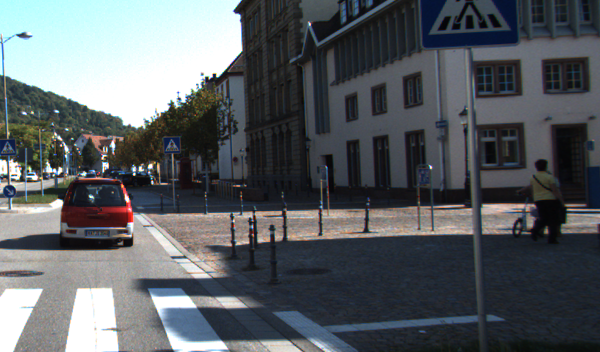} & \includegraphics[width=0.23\textwidth]{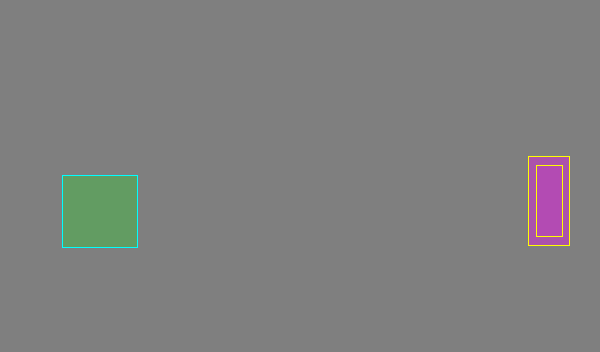} & \includegraphics[width=0.23\textwidth]{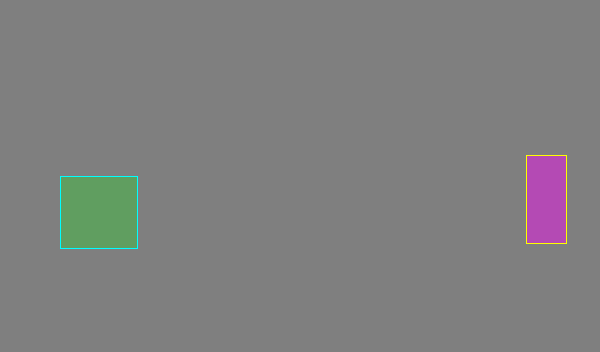} &
     \includegraphics[width=0.137\textwidth]{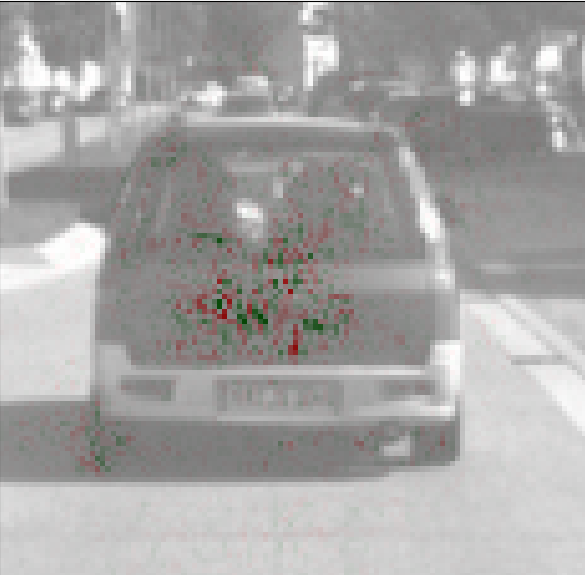}  &
     \includegraphics[width=0.137\textwidth]{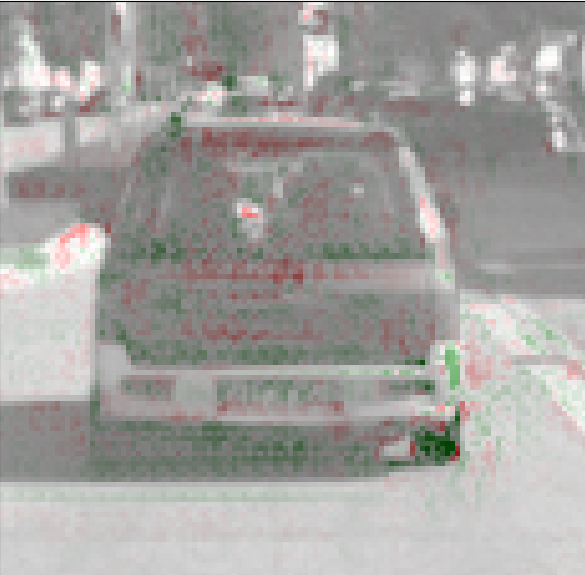}
     \\

     \raisebox{2.6em}{\rotatebox[origin=t]{90}{{\small \textit{Moving Fast}}}}& \includegraphics[width=0.23\textwidth]{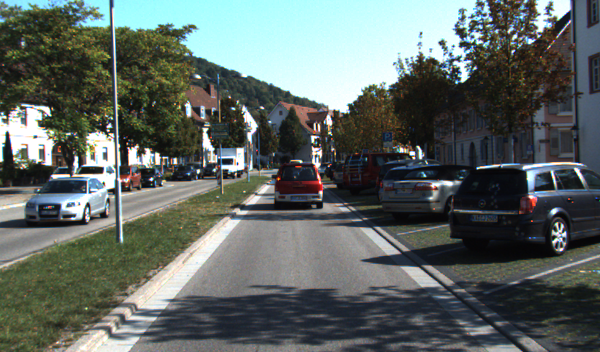} & \includegraphics[width=0.23\textwidth]{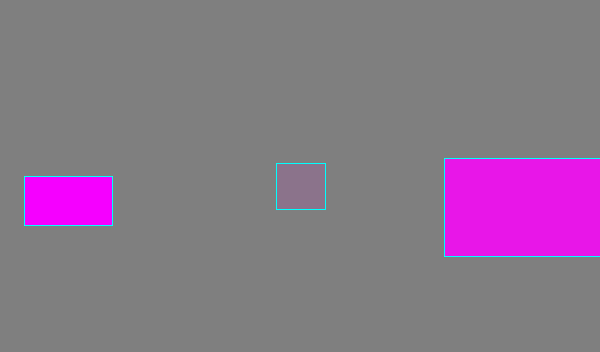} & \includegraphics[width=0.23\textwidth]{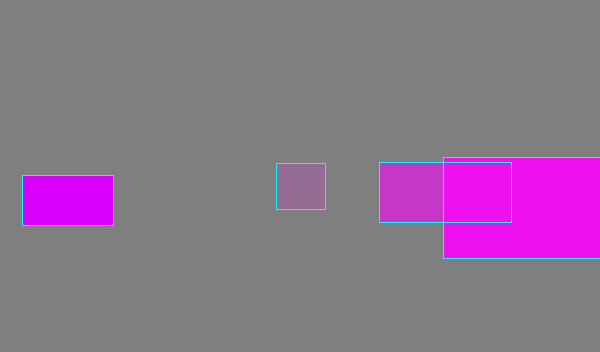} &
     \includegraphics[width=0.137\textwidth]{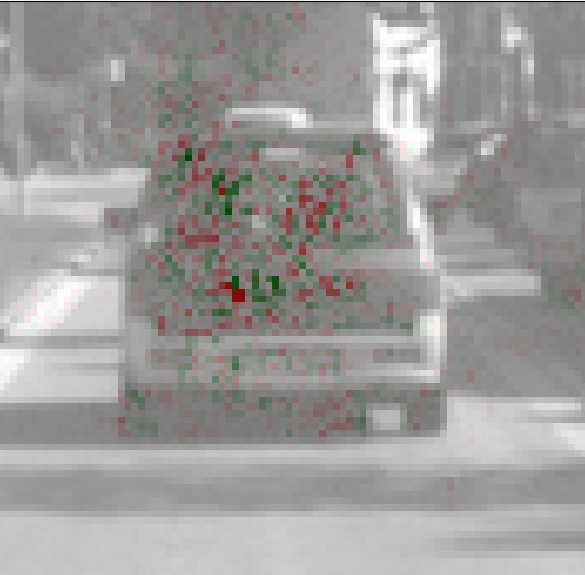}  &
     \includegraphics[width=0.137\textwidth]{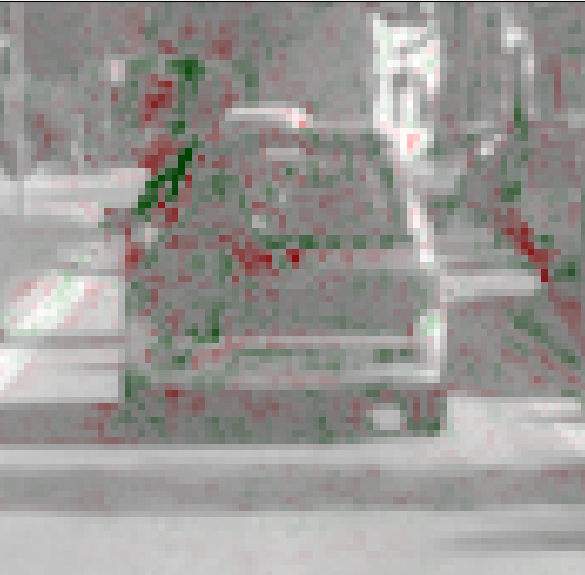} \\

     \raisebox{2.6em}{\rotatebox[origin=t]{90}{{\small \textit{Braking}}}}& \includegraphics[width=0.23\textwidth]{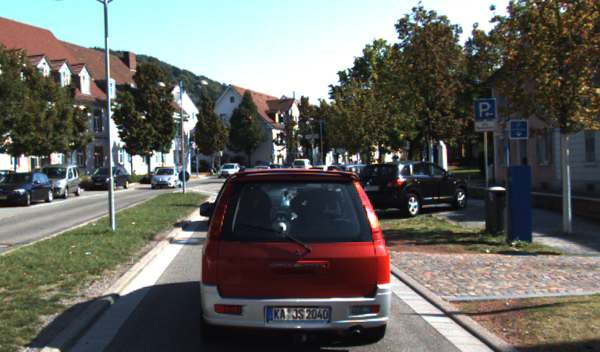} & \includegraphics[width=0.23\textwidth]{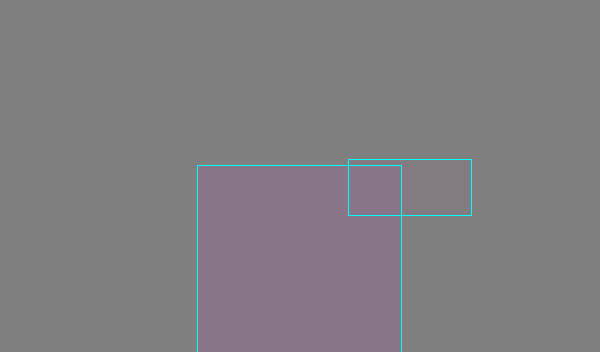} & \includegraphics[width=0.23\textwidth]{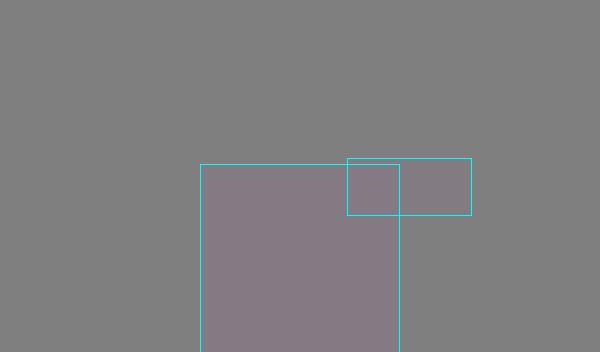} &
     \includegraphics[width=0.137\textwidth]{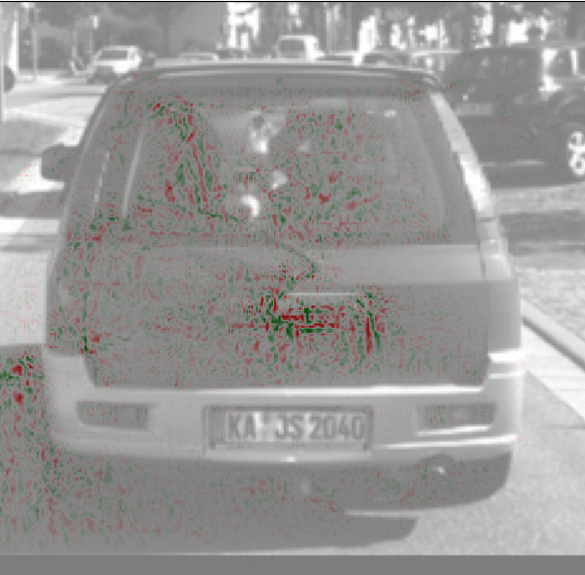}  &
     \includegraphics[width=0.137\textwidth]{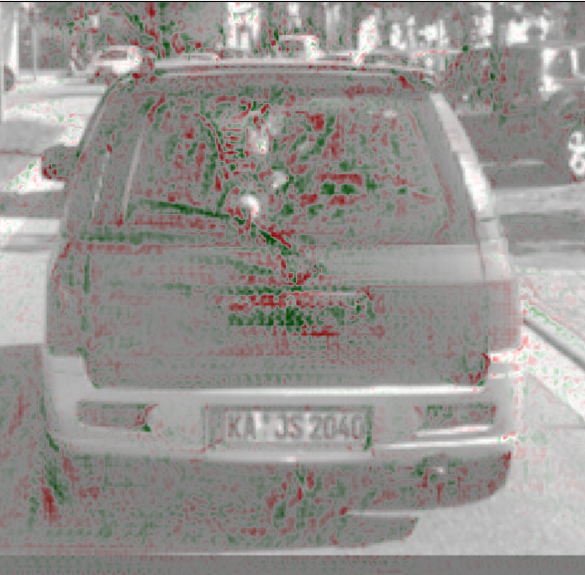} \\

     \raisebox{2.6em}{\rotatebox[origin=t]{90}{{\small \textit{Moving Slow}}}}& \includegraphics[width=0.23\textwidth]{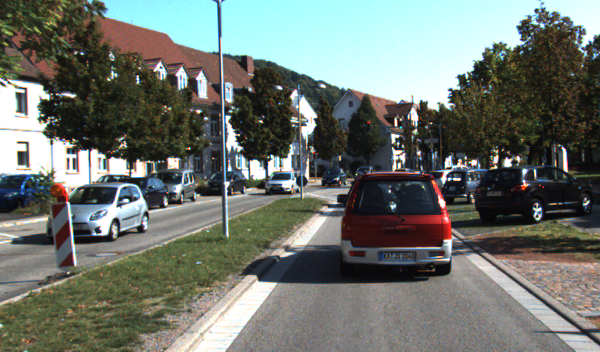} & \includegraphics[width=0.23\textwidth]{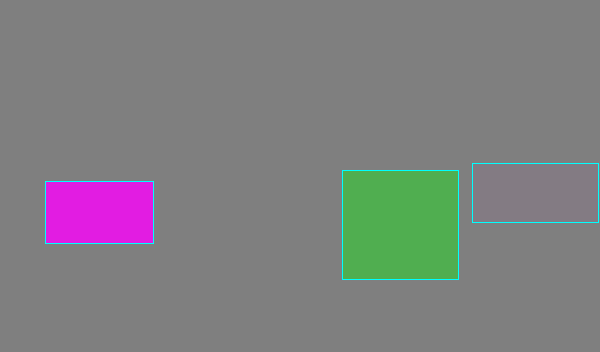} & \includegraphics[width=0.23\textwidth]{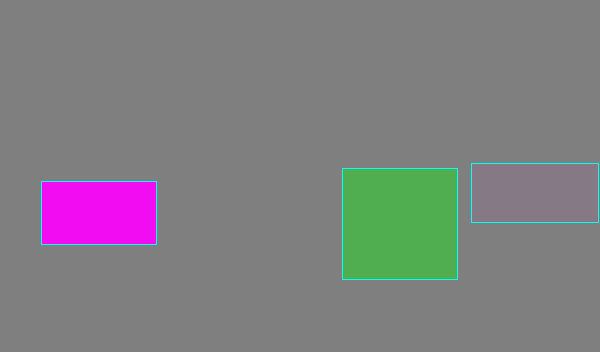} &
     \includegraphics[width=0.137\textwidth]{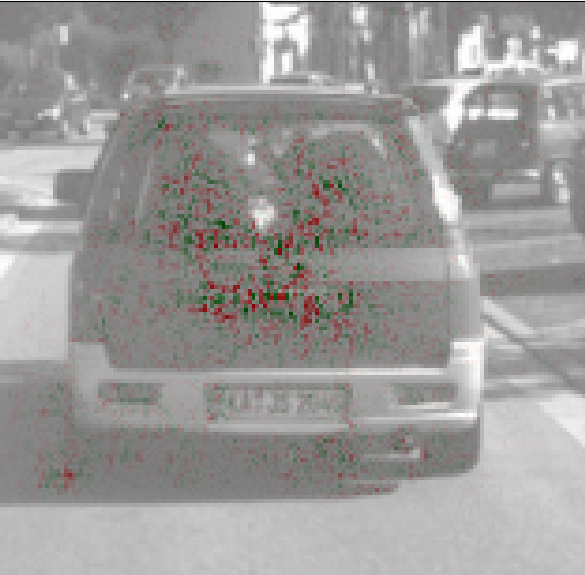}  &
     \includegraphics[width=0.137\textwidth]{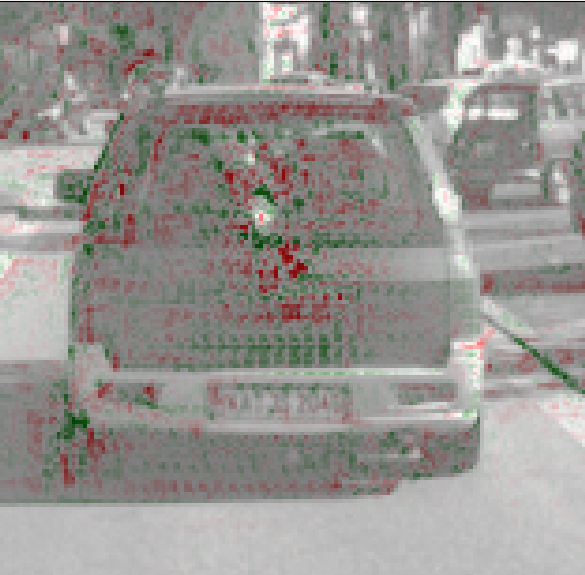} \\

     \raisebox{2.6em}{\rotatebox[origin=t]{90}{{\small \textit{Almost Stopped}}}}& \includegraphics[width=0.23\textwidth]{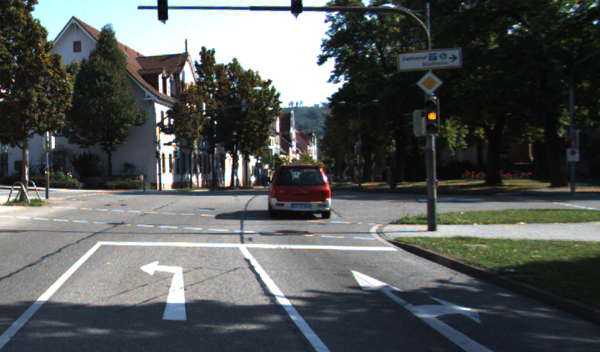} & \includegraphics[width=0.23\textwidth]{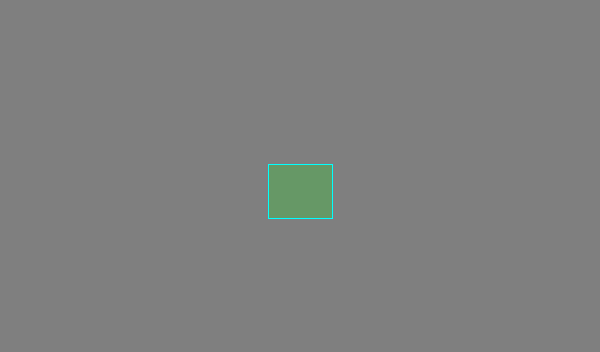} & \includegraphics[width=0.23\textwidth]{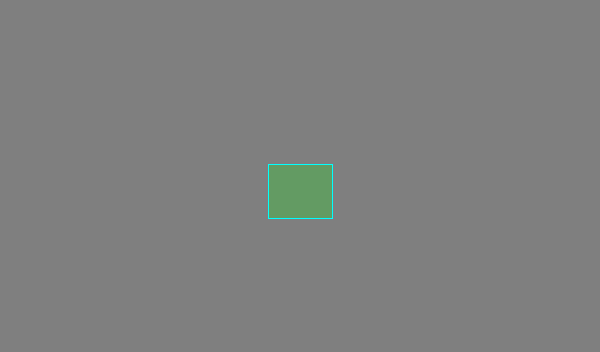} &
     \includegraphics[width=0.137\textwidth]{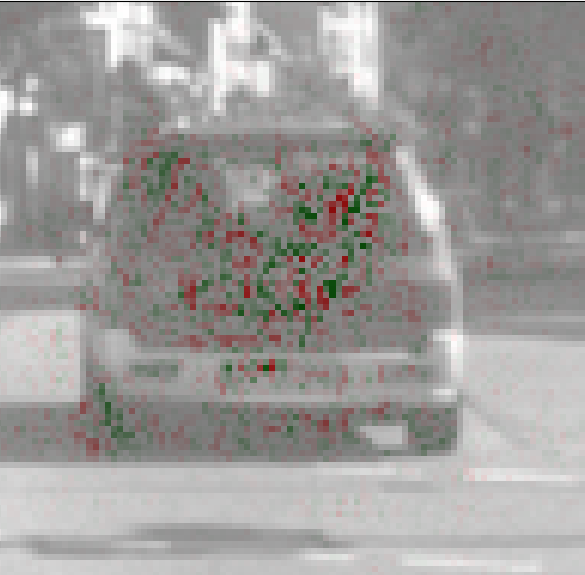}  &
     \includegraphics[width=0.137\textwidth]{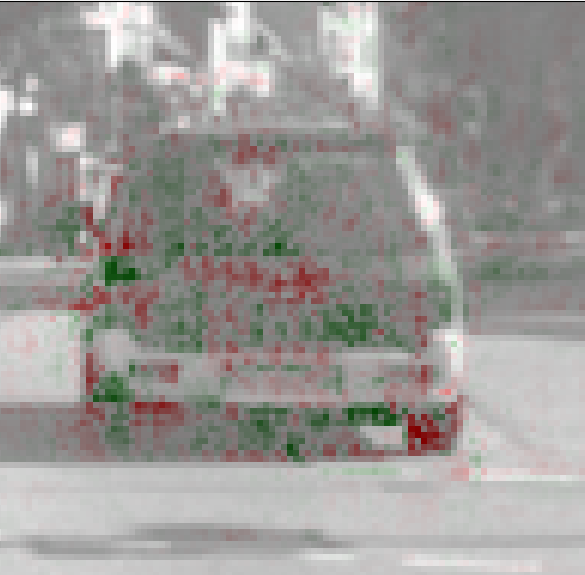} \\

     & & \multicolumn{2}{c}{\includegraphics[width=0.462\textwidth]{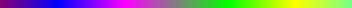}} & \multicolumn{2}{c}{\includegraphics[width=0.277\textwidth]{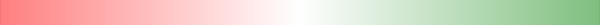}} \\

     & & $\leftarrow$ \textit{\small moving towards} & \textit{\small moving away} $\rightarrow$ &  $\leftarrow$ \textit{\small negative} & \textit{\small positive} $\rightarrow$
     
\end{tabular}

\caption{Shows qualitative comparison and explanations for oTTC predictions across a range of scenarios. The integrated gradients \cite{sundararajan2017axiomatic} method was used to generate explanations. The input and output images are cropped for better visibility. The blue outline in GT and oTTC indicates the cars, while the yellow outline indicates the pedestrians.
}
\label{fig:qual_ig}
\end{figure*}

\subsection{oTTC vs Depth and Velocity}
In addition to 2D location of the objects, the monocular 3D object detectors aim to predict depth information. The monocular 3D object detectors like FCOS3D additionally predict the velocity, orientation and 3D dimensions of the object \cite{wang2021fcos3d}. The velocity and depth information can be used to calculate TTC directly, however, the accuracy of TTC calculated this way can be lower since it is calculated using two predicted values. To verify this, we compare our oTTC model with FCOS3D by calculating TTC using velocity and depth predicted by FCOS3D \cite{wang2021fcos3d}. For this experiment, we use NuScenes dataset \cite{caesar2020nuscenes}.

Table \ref{tab:mono3d_nu} shows the comparison of FCOS3D \cite{wang2021fcos3d} and oTTC. It is evident that the overall performance of oTTC is better compared to FCOS3D in terms of $oMiD$. Moreover, it is important to note that oTTC performs better for predicting TTC of relatively dynamic objects like cars and motorcycles, while FCOS3D performs better for relatively static objects like Pedestrians. This indicates that TTC calculated from velocity and depth is less precise compared to direct TTC prediction. The difference in performance of static and dynamic objects, in case of FCOS3D \cite{wang2021fcos3d}, indicates imprecise velocity predictions. Furthermore, predicting velocity requires velocity ground truth, but time-to-contact can be predicted regardless of velocity annotations.

\subsection{oTTC vs Tracked Depth}
Mono3D object trackers additionally predict depth information per object. This depth information along with object tracks can be used to obtain $MiD$ using Eq. \ref{eq:rod}. For this experiment, we take CenterTrack \cite{zhou2020tracking} as the mono3D tracker and compare it with our oTTC model. Tab. \ref{tab:track_kitti} shows the detailed results of this experiment. It is established that, oTTC performs better than CenterTrack \cite{zhou2020tracking} in terms of $MiD$. As mono depth prediction includes absolute depth scale, it can be tricky to extract this information from a single image. However, relative depth is easier for a network to model given a single frame and the only thing the network needs in the case of Object TTC. Furthermore, in the case of $MiD$ from Mono3D object tracking we combine two values predicted by network with precision $p$, hence the precision of their combination in MiD will be $p^2$.

\subsection{Qualitative Results \& Explanations}
To evaluate our approaches in different scenarios, we conduct a qualitative study across various driving situations including braking, moving and stopping. We also conduct explainability study to explore the contributions of different object features towards the TTC prediction. For these studies, we use the $11^{th}$ video sequence of KITTI tracking \cite{geiger2013vision} dataset, which is also part of the validation set.

Fig. \ref{fig:qual_ig} shows the results of qualitative and explainability study. The $GT$ column represents the ground truth time to contact values per object (oTTC), generated using tracked depth values as explained in Sec. \ref{sec:gt_3dot}. The $oTTC$ column shows the continuous $oTTC$ predictions. The legends are shown at the bottom where the gray color indicates the background or zero velocity relative to the ego vehicle, while green to yellow colors show the object moving away, and pink to blue colors show the object moving toward the ego vehicle. It is evident that the results are similar to ground truth in all scenarios.

The last two columns in Fig. \ref{fig:qual_ig} show explanations of TTC predictions based on integrated gradients \cite{sundararajan2017axiomatic} method. The \textit{Intp. bTTC} shows the explanations of binary risk prediction while \textit{Inp. oTTC} shows explanations of continuous risk estimation for the object under focus. For bTTC model focuses on the object itself with high weight to the area around the center of the object. The bTTC explanations are not conclusive enough to connect the bTTC prediction with high gradient response or motion blur. However, the oTTC explanations clearly have focus on high gradient regions where motion blur effects are maximum, pointing towards the possible high contribution of motion artifacts in the final prediction decision.

%% file: sec/7_conclude.tex
\section{Conclusion}

Time-to-Contact is more useful to model object motion in traffic scenes compared to velocity and depth. The recent research suggests estimating TTC in a per-pixel manner. Such estimation, however, is not useful without object detection, as pixels which belong to the road and are next to the ego vehicle may have least TTC. Also, keeping TTC estimation and object detection as independent tasks yields an inefficient solution. This paper suggests estimating TTC as an object attribute by extending an object detection architecture. By doing so, it yields an efficient solution which provides object detection as well as per object TTC with minimal additional computations. Moreover, compared to previous research, the proposed approach uses only a single image to predict TTC values per object and argues that motion artifacts in images caused by moving camera may be helpful in inferring TTC. To support the argument, this paper conducts a series of experiments, including binary risk estimation and continuous TTC estimation. And compares the TTC results with existing state-of-the-art to prove that the proposed approach has higher precision in terms of MiD.



\section*{Acknowledgments}
This work was supported by the German Ministry for Economic Affairs and Climate Action (BMWK) project KI Wissen under Grant 19A20020G.